# Mental State Recognition via Wearable EEG

Pouya Bashivan[1], Irina Rish[2], Steve Heisig[2]

[1] Computer Vision, Perception, and Image Analysis Lab, The University of Memphis, Memphis, TN, USA
[2] IBM T. J. Watson Research Center, Yorktown Heights, NY, USA
pbshivan@memphis.edu, {rish, heisig}@us.ibm.com

**Abstract**: The increasing quality and affordability of consumer electroencephalogram (EEG) headsets make them attractive for situations where medical grade devices are impractical. Predicting and tracking cognitive states is possible for tasks that were previously not conducive to EEG monitoring. For instance, monitoring operators for states inappropriate to the task (e.g. drowsy drivers), tracking mental health (e.g. anxiety) and productivity (e.g. tiredness) are among possible applications for the technology. Consumer grade EEG headsets are affordable and relatively easy to use, but they lack the resolution and quality of signal that can be achieved using medical grade EEG devices. Thus, the key questions remain: to what extent are wearable EEG devices capable of mental state recognition, and what kind of mental states can be accurately recognized with these devices? In this work, we examined responses to two different types of input: instructional ('logical') versus recreational ('emotional') videos, using a range of machine-learning methods. We tried SVMs, sparse logistic regression, and Deep Belief Networks, to discriminate between the states of mind induced by different types of video input, that can be roughly labeled as 'logical' vs. 'emotional'. Our results demonstrate a significant potential of wearable EEG devices in differentiating cognitive states between situations with large contextual but subtle apparent differences.

## 1 Introduction

Insights about internal states of mind can be a valuable resource in many situations. They can be used to preempt risky situations such as traffic accidents on the roads through monitoring drowsiness in train orbus drivers [4, 5, 9]. EEG headsets along with other wearable devices (e.g., those measuring galvanic skin response and heartbeat rate) can be prescribed to monitor mental states in patients with certain psychiatric conditions such as anxiety or depression [8]. Another possible application involves assessing different reaction types in response to a stimulus. Knowing whether an individual is reacting emotionally or rationally to a speech or advertisement can be used to tailor the content or style and to perform social analysis [11]. Much of the current research on mental state monitoring exploits devices such as MRI scanners or medical grade EEG [7]. Despite their superior signal quality and higher resolution, their usage difficulty has hindered their application in real world situations. On the other hand, consumer grade EEG headsets are affordable and easier to use, but lack the resolution and signal quality of medical grade EEG devices.

Our general objective is to understand to what extent wearable EEG devices are capable of mental state recognition, given their limited number of electrodes and higher levels of noise, and what types of mental states can be accurately detected using such devices. Herein, we focused on a specific study in an attempt to discriminate between different mental states that we roughly denoted as 'rational' vs 'emotional'. We used features extracted from EEG signal recorded using a commercial grade EEG headset with limited number of channels. We designed the experiment to minimize visual and auditory differences between the stimuli representing each state. Particularly, we used two types of input, instructional (Khan academy) versus recreational (cat) videos, in order to produce mental states along the lines of 'rational/logical' versus 'emotional'. Encouraging empirical results (around 75% classification accuracy) were obtained using several machine-learning techniques.

Note that, however, the effect of videos on mental states is much more complex, and can vary depending on the interpretation of the viewer so such distinctions are approximate. Overall, this work should be considered as an initial proof-of-concept study demonstrating promising results when using specific wearable EEG device (InteraXon MUSE) for mental state recognition.

## 2    Methods

### 2.1    Experiment and Device

We collected EEG data from sixteen individuals (6 female), between 25 and 58 years old (41.8±11.2). Participants watched two instructional (youtube Khan Academy) and two recreational (youtube cat) videos in two sessions. Each session contained one video from each category. Each video was around seven minutes long. Participants were instructed to avoid movement and excessive facial muscle movement throughout the experiment. Data for three participants were removed from the dataset due to insufficient amount of clean data remaining after artifact removal procedure. EEG data were collected using a four-channel dry electrode headset (Muse, InteraXon Inc.). Four electrodes were located at standard 10-20 coordinates (T9, FP1, FP2, T10); see http://www.choosemuse.com/what-does-it-measure/ for details. Sampling frequency was 220 Hz which facilitated the extraction of standard EEG bands using Discrete Wavelet Transform (see section 2.3).

### 2.2 Preprocessing

We employed amplitude and variance filters to detect and discard sections of the EEG record containing obvious artifacts. The amplitude of raw EEG was compared against an empirically selected threshold value and a window surrounding the exceeding values (0.5 second pre and 1 second post) were flagged as artifacts. Secondly, variance of EEG was computed for all consecutive one-second windows. Computed variance was also compared against an empirically selected threshold and samples within the window were flagged if the variance levels exceeded the empirically selected threshold. Outputs of both methods were combined into a single measure of artifact which was used to mask out sections during the feature extraction step.

### 2.3 Spectral Power Features

Oscillatory responses are the most common and well-studied characteristics in EEG [1, 2, 10]. In order to extract spectral characteristics we adopted the Daubechies 'db4' discrete wavelet transform. We applied a 6-level DWT on the continuous EEG recording for each video and each channel giving rise to 7 time series of coefficients corresponding to frequency bands in Table 1. We then expanded the coefficients for each level to construct a spectrogram-like representation. Using the mask we derived from artifact detection analysis we discarded the coefficients corresponding to segments of data containing artifacts. We applied the mask to extracted coefficients rather than raw EEG to maintain the continuity of signal which results in more accurate power estimation with DWT. We then computed three types of spectral features. The average and variance of power was computed for each frequency band and for each electrode (average power and power variance). Finally, the hemispheric difference in observed power within each frequency band was computed.

**Table 1.** Corresponding frequency bands and their classic names for each DWT level

| DWT Level | Frequency Range (Hz) | Corresponding Band Name |
|---|---|---|
| 1 | 110-220 | HH-Gamma |
| 2 | 55-110 | H-Gamma |
| 3 | 27.5-55 | L-Gamma |
| 4 | 13.75-27.5 | Beta |
| 5 | 6.88-13.75 | Alpha |
| 6 | 3.44-6.88 | Theta |
| 7 | 0-3.44 | Delta |

### 2.4 Connectivity Features

While features such as spectral power have been used in numerous other studies, there is growing evidence that the connection strength in neural networks is an informative marker [6, 12]. We computed the pairwise correlation between the power within each frequency band and electrode. While correlation is one of the simplest connectivity measures, it is nonetheless suitable for our dataset because of its applicability to discontinued time series (caused by artifacts throughout the recording). By using power correlations, we captured cross-region, cross-frequency interactions in cortical activity visible in EEG recording.

### 2.5 Classifying mental states

We employed a range of linear and nonlinear classifiers on our dataset to explore the data from different angles. To extract samples from continuous EEG we applied a rolling window. For this, we selected a window of DWT coefficients, removed parts corresponding to artifacts and extracted the feature values from the remaining coefficients within the window. The next sample was derived by rolling the window half the window size. We investigated various window sizes including 5, 10, 30, 60, and 120 seconds. For spectral power features, we averaged DWT coefficients for each band across all valid coefficients. For connectivity features, we computed pairwise correlation for the valid data within each window.

Our classification task was predicting the class label (video type) from samples of EEG recorded while participants viewed the video segments. We tried two approaches to learning EEG patterns for each video type. In the first approach (intra-subject), the model was trained and tested on data from only one participant. A two-fold cross validation was performed on the data from the two sessions. In the second approach (inter-subject), we followed the leave-subject-out cross validation approach (13-folds) in which we trained the model on data from all participants except one that was assigned for testing. Details for training each of the classifiers used in our study are summarized below.

**Logistic Regression**
L1-regularized logistic regression is a binary linear classifier that given a set of predictors ($x_i \in \Re^n$) and labels ($y_i \in \{-1,1\}$) solves the following unconstrained optimization problem:

$$\min_w \|w\|_1 + C \sum_{i=1}^{l} \log(1 + e^{-y_i w^T x_i}) \qquad (1)$$

in which $w$ is the vector of weights and $C > 0$ is a penalty parameter. The L1-regularization generates a sparse solution $w$. Optimal parameter $C$ was selected through cross-validation on training data in which the logarithmic range of $[10^{-2} - 10^3]$ was searched.

**Support-Vector Machines**
Support-Vector Machines (SVM) operate by minimizing the loss function:

$$\min_w \frac{1}{2} w^T w + C \sum_{i=1}^{l} \max(0, 1 - y_i w^T \phi(x_i)) \qquad (2)$$

where *w* is the vector of weights, *C* > 0 is a penalty parameter and $\phi(x)$ is a kernel function applied on the input data which was chosen as radial basis function here. SVM hyper parameters consisting of regularization penalty parameter (C) and inverse of RBF kernel's standard deviation ($\gamma = 1/\sigma$) were selected by grid-search through cross-validation (C = {0.01, 0.1, 1, 10, 100}, $\gamma$ = {0.1, 0.2, …, 1, 2, …, 10}).

**Deep Belief Network**

We used a three-layer Deep Belief Network (DBN). The first layer was a Gaussian-Binary Restricted Boltzman Machine (RBM) and the other two layers were Binary RBM. The output of the final level was fed to a two-way softmax layer for predicting the class label. Parameters of each layer of DBN were greedily pre-trained to improve learning by shifting the initial random parameter values toward a good local minimum [3]. Number of neurons in the three layers were empirically selected as 1000, 500, and 100. Network was fine-tuned using batch stochastic gradient descent with L1-regularization to reduce the overfitting effect during training.

**Random Forest**

Random forest is an ensemble method consisting of a group of independent random trees. Each tree is grown using a subset of features randomly selected. For each input, outputs of all trees are computed and the class with majority of votes is selected. Number of estimators for the random forest was varied between 5 and 20.

## 3      Results

In the first approach we performed a 2-fold cross validation on data for each individual (intra-subject). The positive label was assigned to the cat video class. The false-negative rate (FNR), false-positive rate (FPR) and combined error rate was averaged over all folds and participants. Window sizes ranging from 5 to 120 seconds were evaluated with SVM, Logistic Regression, and Random forest classifiers (Fig. 1). Among the three selected classifiers, the best result was achieved using SVM model. While performance of LR and RF remained almost the same with different window sizes, both classification errors drastically decreased for SVM by increasing the window duration used for sampling from EEG recording (FPR: 31.5%, FNR: 19.8%, Combined: 25.5%). The best result was achieved for highest window size (120 seconds).

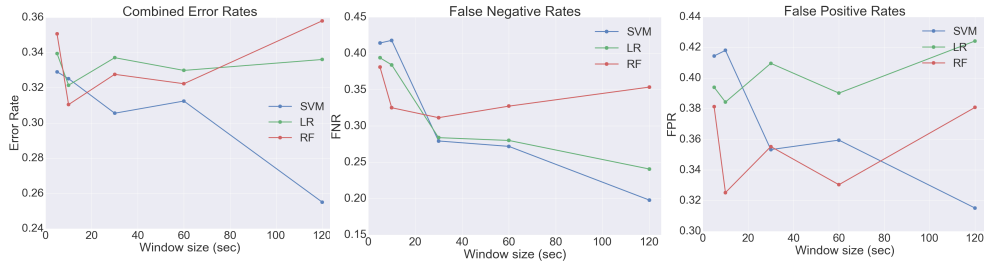

**Fig. 1.** Combined error rates, FNR, and FPR for intra-subject classification.

In the second approach, we performed a single 13-fold (leave-subject-out) cross validation and averaged the error rates over all folds (inter-subject). Here, LR and DBN performed best among the four selected classifiers (Fig. 2). However, looking at the two error rates (FNR and FPR) we found that the two error rates were highly unbalanced for the DBN classifier (FNR= 6.2%, FPR=38.3%), giving preference to categorizing data as cat videos. While LR had a slightly higher combined error, Type I and II errors were much more balanced (FPR: 27.9%, FNR: 23.3%, Combined: 26%).

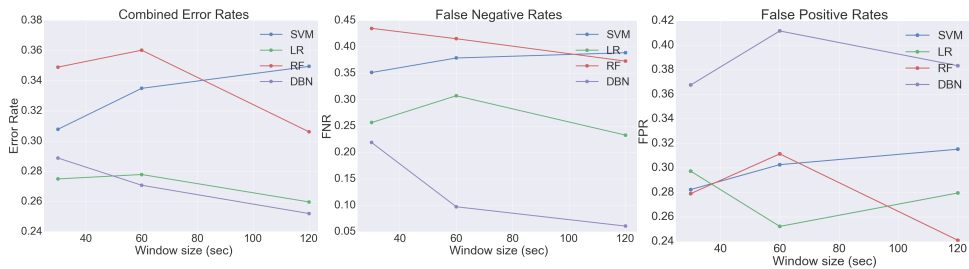

**Fig. 2.** Combined error rates, FNR, and FPR for inter-subject classification.

Because of its inherent feature selection characteristic due to L1-regularization, the trained logistic regression model can provide intuition about the critical features that determine the differences between mental states (classes). The fully trained LR model had a total of 31 non-zero weights corresponding to 7% of all features. This configuration was obtained for almost the sparsest solution. Among these 31 features, none belonged to average power category, one was from hemispheric power difference, six belonged to variation in power, and 24 belonged to correlation features (Fig.3).

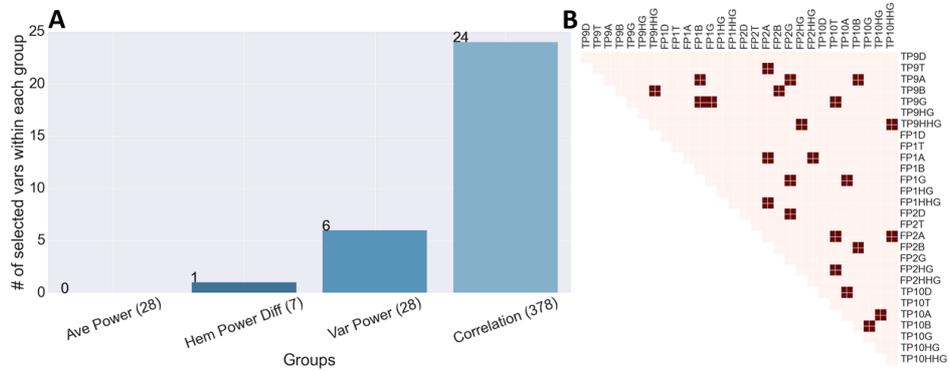

**Fig. 3.** (A) Categorical distribution of L1-logistic regression selected features (C=0.1), (B) Map of selected correlation features.

Using our best classification results with intra and intra subject approaches, we evaluated the "predictability" of each of participants. We used the TP and TN rates for each individual (Fig. 4). Comparing the results for the two approaches, the intra-subject model demonstrates a very unbalanced choice of class. For 11 of the individuals the error rate for one of the classes was zero and the amount of error gained for the second class determined the overall performance. However, for the inter-subject case, the errors are much more balanced. By incorporating data from multiple individuals we included more of the variability in the input space to train the model which led to a more generalizable model. Among all participants, seven individuals had both TPR and TNR greater than 60%.

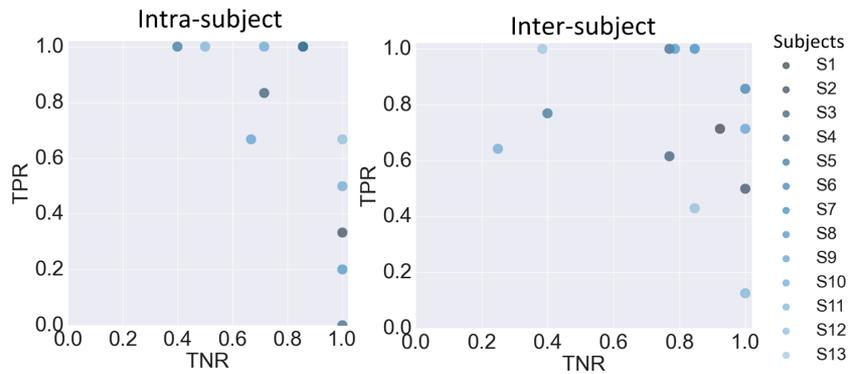

**Fig. 4.** Individual predictability.

## 4  Discussion, Issues and Open Questions

Despite the limited number of electrodes available in the Muse device (4 electrodes covering prefrontal and occipital/temporal regions), we found that a model can be trained to distinguish between two mental states with abstract contextual differences. The window length used to generate samples from continuous EEG had a decisive effect on the achievable performance of the trained model. Samples generated from longer time windows tended to grasp more temporally invariant measurements. The 120 second window for which our best results were seen might be too long for some applications. However, even higher window sizes may still be used in applications with the goal of long term tracking.

Looking deeper at the features automatically selected by the LR model, it is noteworthy that none of the average power features which are the most common features were selected by the LR model. Instead we observed that pairwise connectivity measures among power measures as well as variations in power were the preferred choices. This observation is corroborated by recent findings on importance of cross-regional interconnections in various neuroimaging modalities [12].

It is important to note that while evaluating the wearable device, we discovered many sources of artifacts during our experiments, though many of them are shared with medical grade EEG. The largest class, as expected, was muscle artifacts. Since brain signals are on the order of microvolts and muscle signals in the millivolt range any muscle movements in the face or scalp are potentially problematic. Blinking and jaw clenching events are noted by the device and are usually fairly short duration. Movements of the corregator supercilii, orbicularis oculi, and frontalis muscles caused by squinting or emotional reactions are longer lasting and were more difficult to detect. It was easier to get a good contact at the temporal points without glasses, so squinting was more of an issue that might normally be expected. Another type of muscle issue was glosso-kinetic effects caused by movement of the tongue since the tip is negatively charged and the root is positive. Electro-ocular effects caused by movements of the eyeballs were a constant problem. The cornea is positively charged and the retina is negatively charged so the eyeballs are dipoles and any movement produces artifacts. We observed a subject with nystagmus whose involuntary eye movements produced an enormous amount of delta noise. Even during closed eye experiments some subjects exhibited blepharospasms (eyelid fluttering) which produce artifacts. These effects are all present in medical grade devices as well. Wearable specific effects included salt bridges and electrode popping where a proper contact is not achieved between skin and electrode. A moderate amount of perspiration can cause a slow change in conduction which manifests as a very slow wave, while a larger amount can cause a total disconnection of the electrode. It was common for inexperienced users to wear the device too low on the forehead so the electrodes were placed over a sinus cavity rather than brain. This led to weaker than expected reading for frontal signals. Users with thick hair typically had problems getting a good contact with the temporal electrodes. Prescription medications and illicit drugs have response stereotypes need to be taken into account, and the interactions between multiple medications are particularly intractable. We also found that experienced meditation practi-tioners have a very noticeable profile. Ultimately, in our experience, it was worth every effort to collect the best data possible rather than either discarding data with artifacts or filtering around it some other way during analysis.

Overall, the device we tested was only wearable for a fairly short amount of time. It got uncomfortable after more than 30 minutes or so. This makes it only useable for session-based exercises. The tradeoff of dry electrodes over gel-based electrodes was another wearability issue, since dry electrodes are more prone to losing contact. Most of the subject-based artifacts we observed would also appear in medical grade devices, but scenarios such as driving that are very good fit for wearable devices induce more of them. The limited number of electrodes versus medical grade EEG devices is also a restriction especially when discarding data for artifact reasons. However, we believe that with the development of novel approaches, such as thin tattoo-type electrodes, EEG recording will become practical in a much wider spectrum of real-life scenarios, and that recognizing mental states using wearables devices will become much more widespread. A widely open direction for further research is therefore to continue exploring the potential of wearable EEG for mental state recognition, discovering which types of mental states tend to be easy or hard to recognize, and what types of features extracted from raw data are best for characterizing those states.